\documentclass[11pt,a4paper]{article}
\usepackage[hyperref]{acl2021}
\usepackage{times}
\usepackage{latexsym}

\usepackage{booktabs}
\usepackage{multirow}
\usepackage{amsmath, amsfonts}
\usepackage{pifont}

\usepackage{graphicx}

\usepackage[T1]{fontenc}

\usepackage[utf8]{inputenc}

\usepackage{microtype}

\usepackage{hyperref}
\usepackage{booktabs}
\usepackage{graphicx} 
\usepackage{subcaption}
\usepackage{xspace}

\usepackage{ifthen}

\usepackage{cuted}

\usepackage{makecell}
\usepackage{tabularx}

\usepackage{multirow}
\usepackage{colortbl}

\usepackage{amsmath,amsfonts,bm}

\def\eqref#1{equation~\ref{#1}}

\def\1{\bm{1}}

\def\vc{{\bm{c}}}

\def\ve{{\bm{e}}}

\def\vh{{\bm{h}}}

\def\vx{{\bm{x}}}
\def\vy{{\bm{y}}}

\DeclareMathAlphabet{\mathsfit}{\encodingdefault}{\sfdefault}{m}{sl}
\SetMathAlphabet{\mathsfit}{bold}{\encodingdefault}{\sfdefault}{bx}{n}

\aclfinalcopy 

\makeatletter
\renewcommand*{\@fnsymbol}[1]{\ensuremath{\ifcase#1\or \mathsection\or \dagger\or \ddagger\or
    \mathsection\or \mathparagraph\or \|\or **\or \dagger\dagger
    \or \ddagger\ddagger \else\@ctrerr\fi}}
\makeatother

\title{AgreeSum: Agreement-Oriented Multi-Document Summarization}

\author{Richard Yuanzhe Pang$^{1*}$\thanks{\hspace{2.2mm}Work completed during internship at Google Research.}~~~~~~~Adam D. Lelkes$^{2*}$~~~~~~~Vinh Q. Tran$^{2*}$~~~~~~~Cong Yu$^2$ \\
$^1 $New York University, New York, NY 10011, USA \\
$^2 $Google Research, New York, NY 10011, USA \\
{\tt yzpang@nyu.edu, \{lelkes,vqtran,congyu\}@google.com}}

\date{}

\begin{document}
\maketitle

{\let\thefootnote\relax\footnotetext{$^{*}$~Equal contribution.}}

\begin{abstract}
We aim to renew interest in a particular multi-document summarization (MDS) task which we call AgreeSum: agreement-oriented multi-document summarization. Given a cluster of articles, the goal is to provide abstractive summaries that represent information common and faithful to all input articles. Given the lack of existing datasets, we create a dataset for AgreeSum, and provide annotations on article-summary entailment relations for a subset of the clusters in the dataset. We aim to create strong baselines for the task by applying the top-performing pretrained single-document summarization model PEGASUS onto AgreeSum, leveraging both annotated clusters by supervised losses, and unannotated clusters by T5-based entailment-related and language-related losses. Compared to other baselines, both automatic evaluation and human evaluation show better article-summary and cluster-summary entailment in generated summaries. On a separate note, we hope that our article-summary entailment annotations contribute to the community's effort in improving abstractive summarization faithfulness. 
\end{abstract}

\section{Introduction}
\label{sec:intro}

Recent works have made great progress in single-document summarization (SDS) thanks to the encoder-decoder framework and pretraining procedures~\citep{cho-etal-2014-learning,rush-etal-2015-neural,narayan-etal-2018-dont,zhang2020pegasus}. There is a growing interest in multi-document summarization \citep[MDS;][et seq.]{zopf-2018-auto,fabbri-etal-2019-multi,liu-lapata-2019-hierarchical,chu2019meansum,wang-etal-2020-heterogeneous}, with applications in search engines, news clustering, timeline generation, and other areas. Past MDS research has primarily focused on summarizing articles such that the summary covers an event ``comprehensively'' while ``avoiding redundancy'' \citep{fabbri-etal-2019-multi}. We can say that most existing MDS tasks summarize the ``union'' of the articles. 

In this paper, we discuss \emph{agreement-oriented multi-document summarization} (AgreeSum), in which we aim to \emph{abstractively} summarize the ``intersection'' of the articles. More specifically, the input to the task is a cluster of articles, and the expected output is a summary that represents information common and faithful to all input articles in the cluster (Section \ref{sec:dataset}). A few works (discussed in Section~\ref{sec:related}) have investigated the problem, without using modern neural-network-based methods. The motivation for reviving interest in AgreeSum is twofold. First, given that certain microscopic details are not likely present in all articles in a given cluster, they would be filtered out through AgreeSum. If the source articles reflect different points of view, AgreeSum provides a way of capturing the common ground.

The second motivation for AgreeSum lies in the pursuit of summarization faithfulness. AgreeSum is timely, given that recent works have shown difficulty of producing faithful abstractive summaries  \citep{falke-etal-2019-ranking,maynez-etal-2020-faithfulness,kryscinski-etal-2020-evaluating,durmus-etal-2020-feqa,zhou2020detecting}, though in the SDS setting. AgreeSum could allow practitioners of abstractive summarization systems to carefully study topics related to faithfulness and hallucination.

Given the scarcity of readily available data, we create a dataset\footnote{\url{https://github.com/google-research-datasets/AgreeSum}} based on English Wikipedia current events portal (WCEP)\footnote{\url{https://en.wikipedia.org/wiki/Portal:Current_events}}. WCEP contains neutral human-written news summaries with links to news articles (with usually one link per summary). We first extract human-written summaries on WCEP, the linked articles, and semantically similar articles to the linked articles, in order to obtain article clusters associated with summaries. We then annotate the entailment relationship for a subset of the cluster-summary pairs (i.e., whether or not a given article from the cluster semantically entails the summary). From there, we build cluster-summary pairs for AgreeSum (Section~\ref{sec:agreesum-dataset}).

We build upon previous SDS research, using the top-performing pretrained PEGASUS model~\citep{zhang2020pegasus} as the starting point for our models. Using our dataset, we first examine a few baseline models and show that several pretraining-based summarization models fail at generating summaries that satisfy AgreeSum's requirements. We also propose an approach that integrates both supervised (using the annotated portion of the dataset) and unsupervised losses (namely, an entailment loss and a language loss to be discussed later, using both the annotated portion and the unannotated portion) while leveraging PEGASUS. 
We show the effectiveness of a simple policy gradient-based algorithm \citep{sutton2000policy} in which the rewards are based on a T5-based article-summary entailment model \citep{raffel2020exploring}. 
To summarize the contributions:
\begin{itemize}
    \item We introduce the AgreeSum task and the WCEP-based dataset (\S\ref{sec:agreesum-dataset}). 
    \item A subset of \emph{article}-summary pairs are annotated with entailment information (\S\ref{sec:entailment-dataset}). On a separate note, the article-summary-level recognizing textual entailment (RTE)\footnote{The RTE datasets \citep{dagan2005pascal, haim2006second, giampiccolo-etal-2007-third} correspond to the following two-class classification task: given a premise and a hypothesis, a model would decide if the premise entails the hypothesis.} task could stand as a challenging task on its own. The annotations could be of interest to the research in improving abstract summarization faithfulness, in the context of not only AgreeSum but also general single-document summarization tasks.
    \item We develop simple ways of applying PEGASUS to AgreeSum. We provide a few baselines as well as a model that uses unsupervised entailment-related and language-related losses to complement the supervised finetuning of PEGASUS (\S\ref{sec:methods}). Both automatic and human evaluations are performed (\S\ref{sec:results}). In particular, we show that the T5-based entailment model can be integrated into summarization models to produce summaries that are entailed in source articles. 
\end{itemize}

\section{Related Work}
\label{sec:related}

\paragraph{Traditional MDS.} 

DUC \citep{paul2004introduction,dang2005overview} and TAC \citep{owczarzak2011overview} 
MDS data are among the first high-quality relevant datasets. These datasets are human-curated, but tiny in terms of the number of examples. 
Recent works have explored creative methods for obtaining low-cost MDS datasets. \citet{liu2018generating} use Wikipedia articles as summaries and the cited articles as inputs. \citet{antognini-faltings-2020-gamewikisum} use Wikipedia in a similar way, in the video games domain. \citet{fabbri-etal-2019-multi} rely on the website Newser with lengthy human-aggregated extractive summaries. \citet{gholipour-ghalandari-etal-2020-large}, also based on WCEP, is especially relevant.

However, our dataset is different in the following ways. First, all of our articles in the same cluster are about the same event. Next, a large part of our dataset is annotated with article-summary entailment information (i.e., in each of the clusters, for each article in the cluster, whether the article entails the summary; see Section~\ref{sec:dataset}). Further, among the annotated article-summary pairs, about half of the articles entail the summary, and half of the articles do not entail the summary. This property makes a realistic and difficult setting for AgreeSum tasks. 

In terms of recent MDS neural methods, \citet{chu2019meansum} summarize opinions using an auto-encoder, in which case the input is much shorter than a typical article. \citet{liu2019summae} improve the model by encoding articles and summaries in the same space. Other novel approaches include using sentence compression in the seq2seq framework 
\citep{baziotis-etal-2019-seq}, jointly learning sentence fusion and paraphrasing \citep{nayeem-etal-2018-abstractive}, using graph neural networks to help extraction \citep{wang-etal-2020-heterogeneous}, using spectral methods \citep{wang-etal-2020-spectral}, using transfer learning based on a novel pretraining method called gap-sentence prediction \citep{zhang2020pegasus} on a news-specific corpus, among a few others \citep{li-etal-2020-leveraging,gu2020generating,mao-etal-2020-multi}.

For MDS tasks that summarize the intersection of articles, a few past works have discussed the helpfulness of models that identify common information ``centroids'' among multiple related documents, so as to allow internet users to more efficiently understand events \citep{radev-etal-2000-centroid,barzilay-mckeown-2005-sentence}. The attempted non-neural models rely heavily on topic/theme detection and tracking, and are more extractive than abstractive \citep{radev2004centroid}. The AgreeSum idea has not been fully explored by researchers since much stronger text generation technologies became available. It is timely to revisit the problem also because recent stronger neural abstractive summarization models are prone to hallucination, as are neural text generation in general \citep{wiseman-etal-2017-challenges,tian2019sticking,wang-sennrich-2020-exposure,pang2020text}.

\paragraph{Summarization hallucination and evaluation.} 

Non-hallucination is a necessary but not sufficient condition for performing well in AgreeSum: the summary not only needs to be entailed in the union of the articles, but also must be entailed in each of the articles. Unfortunately, recent works have shown the difficulty of identifying and mitigating hallucination \citep{maynez-etal-2020-faithfulness}. 

Evaluation-wise, researchers have found that metrics like ROUGE \citep{lin-2004-rouge} and BERTScore \citep{zhang2020bertscore} are only weakly correlated with factuality. \citet{durmus-etal-2020-feqa} and \citet{wang-etal-2020-asking} have therefore proposed using question-answering systems for evaluating summarizers. Recently, \citet{zhou2020detecting} have made progress in creating token-level hallucination detectors which rely on negative (i.e., hallucination) data augmentation to train. 

In terms of improving faithfulness and factuality, researchers did not find natural language inference (NLI)\footnote{The NLI datasets, beginning with SNLI \citep{bowman-etal-2015-large}, correspond to a three-class (entailment, neutral, contradiction) entailment classification task.} models trained on standard NLI datasets to be robust enough for summarization-related downstream tasks \citep{falke-etal-2019-ranking}. Contemporaneously, entity chains are used to explicitly ground the generations so that they become more faithful \citep{narayan2021planning}. More broadly, there have been other recent works striving to develop techniques for high-precision generation \citep{malmi-etal-2019-encode,tian2019sticking,pang2020text,parikh-etal-2020-totto,dusek-kasner-2020-evaluating}.

\section{AgreeSum Task and Datasets}
\label{sec:dataset}

\subsection{Task}

\paragraph{Short description of AgreeSum.}

The input is a cluster of around four articles (refer to Section~\ref{sec:agreesum-dataset} for more details) that describe the same event.  
However, the articles may have different levels of details and/or different levels of neutrality; e.g., one article in a cluster may be an opinion, while other articles may be neutral news. The expected output is an abstractive summary that represents information common and faithful to all input articles in the cluster. Moreover, the summary needs to be informative. 

\subsection{English AgreeSum Dataset Based on WCEP}
\label{sec:agreesum-dataset}

\paragraph{Step 1.} Recall that WCEP contains neutral human-written news summaries, each of which is linked to a news article. The first step is to obtain the summaries and 8 on-topic articles\footnote{A small amount of clusters (233, or $\sim$4\%) have fewer than 8 articles to ensure relevance of all articles in a cluster.} for each summary based on WCEP. 
Specifically, we collect 5564 human-written summaries $\{\vy_i\}_{i=1}^{5564}$ from WCEP. 
For each summary, we have one linked article to each summary on WCEP; we call the set of such articles $\{\vx_{i}^{(0)}\}_{i=1}^{5564}$. Given that we want to generate abstractive summaries, and to make the dataset challenging, the set of articles $\{\vx_{i}^{(0)}\}_{i=1}^{5564}$ will \emph{not} be used in the final dataset, to prevent excessive textual overlaps between the input articles and the target summaries. 
For each $i$, we obtain 8 other news articles $\vx_{i}^{(1)}, \vx_{i}^{(2)}, \dots, \vx_{i}^{(8)}$ that are semantically similar to $\vx_{i}^{(0)}$, based on a proprietary BERT-based clustering algorithm.

\paragraph{Step 2.} The second step is to annotate entailment relations. Annotators are asked to judge if each of the articles in a cluster entails the summary (i.e., ``does the article contain all the information presented in the summary?''). 1025 \emph{cluster}-summary pairs are annotated. 
We designate $\sim$10\% of the annotated clusters 
to be in the dev set. This step is discussed further in Section~\ref{sec:entailment-dataset}.

\paragraph{Step 3.} To make AgreeSum moderately but not overly difficult, we set the maximum number of articles per cluster to be four, and the next step is to transform the dataset to have four articles per cluster. 
To take advantage of entailment annotations, we duplicate each annotated cluster in the training set $\vc_i = \{\vx_{i}^{(1)}, \vx_{i}^{(2)}, \dots, \vx_{i}^{(8)} \}$ ten times so that there is roughly an equal number of annotated and unannotated clusters.\footnote{A side effect is that such choices could allow for more supervised cross-entropy updates.} Next, for each cluster, we randomly choose four articles 
to keep in the cluster, and discard the rest.

\begin{figure}[t]
     \centering
         \includegraphics[width=0.30\textwidth]{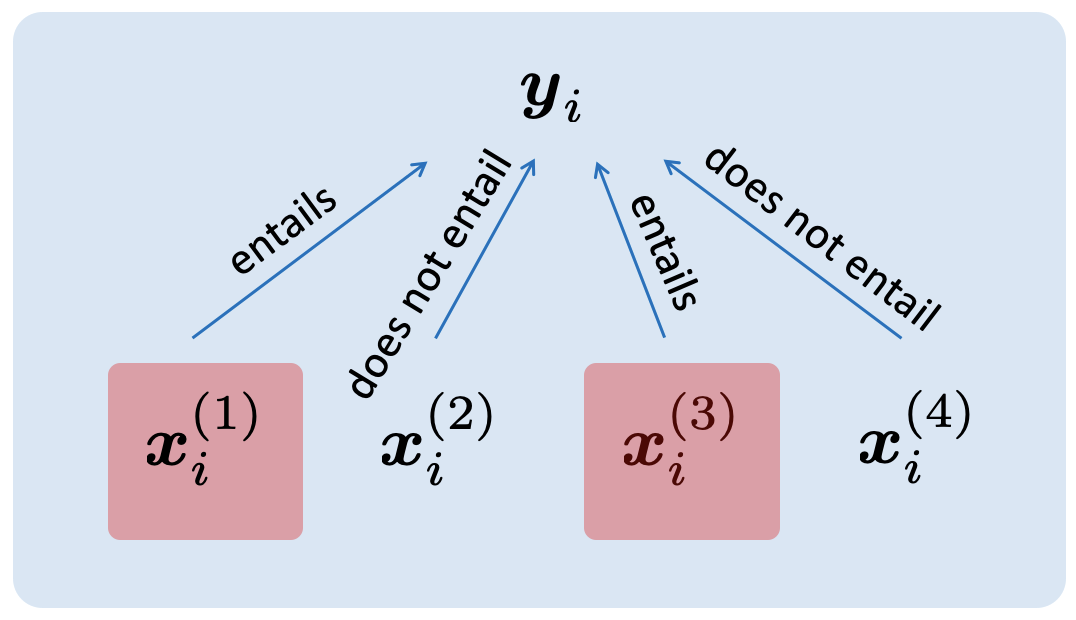}
    \caption{An annotated example (i.e., input-output pair where the input is a cluster of articles and the output is a summary) in the training set. In the figure, two articles in the input cluster entail the summary while the other two do not. An example is either annotated (meaning all article-summary pairs in the cluster-summary pair are annotated) or unannotated (meaning none of the article-summary pairs are annotated). Note that in the dev set, all articles entail the summary given that we would like to compare between generated summaries and the gold summaries.}
    \label{fig:data-example}
\end{figure}

\paragraph{Dev and test sets.} For the development set, we aim to designate the WCEP summaries as gold summaries. Therefore, for each cluster, we only keep the articles that entail the corresponding summary based on human annotation. In the case that a cluster has $>4$ articles that entail the summary, we split the cluster into two, such that each newly formed cluster has $\geq 2$ articles that entail the summary.

We use WCEP entries corresponding to dates before August 2019 for the training and development sets; for the test set, we sample 150 clusters of articles containing WCEP linked articles published between August 2019 to August 2020 to ensure that the test set does not have overlapping articles or overlapping publication times with the training and development sets.

For each sampled test cluster, we sample four articles from the cluster. Unlike the development set, these four articles are not annotated and are not guaranteed to entail any particular common summary. As a result, the provided WCEP summaries in the test set are \textbf{not} gold-standard summaries. 

The test set is expected to be challenging due to construction methods, as well as the shift in topics of recent events (e.g., the COVID-19 pandemic). The test set WCEP summaries are not gold-standard summaries, but are nevertheless provided given the potential use to approximate generation informedness (Section~\ref{sec:results}). 

We thus obtain the dataset split shown in Table~\ref{tab:dataset}.

\begin{table}[h!]
\setlength{\tabcolsep}{3pt}
\centering
\small
\begin{tabular}{lccccccc}
\toprule
& \multicolumn{3}{c}{\makecell[l]{\# of cluster- \\ summary pairs}} & & \multicolumn{3}{c}{\makecell[l]{\# of article- \\ summary pairs}} \\
\cline{2-4} \cline{6-8}
\noalign{\smallskip}
& train & dev & test & & train & dev & test \\
\midrule
all & 18208 & 132 & 150 & & 70137 & 423 & 600 \\
annotated & 9130 & 132 & 0 & & 33841 & 423 & 0 \\
(at least 1) entailed & 7610 & 132 & 0 & & 17951 & 423 & 0 \\
unannotated & 9078 & 0 & 150 & & 36296 & 0 & 600 \\

\bottomrule
\end{tabular}
\caption{Dataset information. A cluster contains $\leq$ 4 articles. The ``(at least 1) entailed'' row refers to the number of \emph{annotated} clusters containing at least one article that entails the summary (in the cluster-summary case), or the number of articles that entail the summary \emph{among the annotated pairs} (in the article-summary case). 
\label{tab:dataset}} 
\end{table}

\section{Article-Summary Entailment}
\label{sec:entailment-dataset}

\subsection{The Entailment Dataset} 

Workers have annotated the entailment relation of 1025 cluster-summary pairs (a subset of the cluster-summary pairs obtained in Step 1 in Section~\ref{sec:agreesum-dataset}), 
which correspond to 7750 article-summary pairs. For each article-summary pair, we ask \emph{five} professionally trained annotators, randomly chosen from a pool of $\sim$800 raters, whether the summary is semantically entailed in the article; we then take the majority answer. See the appendix for more details.

Our AgreeSum models leverage this entailment dataset, as discussed in Section~\ref{sec:methods}. In addition, this dataset could be seen as a challenging RTE-style (or two-class NLI-style) task.

\subsection{Model Performance}

To explore how models perform on the article-summary entailment dataset, and to see whether we can have a good entailment-classification model to guide the learning of our summarizer that encourages faithfulness to all input articles in the same cluster, we treat the dataset as a two-class NLI task where the label is either ``entailed'' or ``not entailed,'' and examine the classification accuracy. Specifically, we evaluated the following three models. RoBERTa-large \citep{liu2019roberta} fine-tuned on MNLI fails on our entailment data\footnote{In this case, the majority of the predictions are ``not entailed.''}, likely in part because the premises in MNLI are sentences but are articles in our article-summary entailment task.  It is also worth noting that unlike T5 \citep{raffel2020exploring}, RoBERTa-large is not pretrained on a multi-task mixture of many tasks, with each task converted into the text-to-text format.

We also attempt a model that integrates the PEGASUS-encoder which is pretrained in the news domain using the HugeNews corpus which contains around 1.5 billion articles. Given that pretrained PEGASUS does not include \texttt{[CLS]} tokens and it is very expensive to re-pretrain, we use a CNN-based classifier~\citep{kim-2014-convolutional} whose input is the PEGASUS-encoder-outputs; more specifically, the convolutional layers pool over the sequence of encoder outputs. Based on the architecture by \citet{kim-2014-convolutional}, our CNNs use filter $n$-gram sizes of 2, 3, 4, and 5, with 256 filters each. The resulting classifier achieves $\sim$68\% accuracy on our entailment dataset.\footnote{Doing intermediate training on MNLI \citep{pruksachatkun-etal-2020-intermediate} and then training on our entailment dataset, unfortunately, does not strengthen performance.}

In comparison, T5~\citep{raffel2020exploring} shows encouraging results. 
The multi-task-trained T5-large fine-tuned on our training set achieves 81.3\% accuracy (79.1\% for vanilla T5-large). 
T5-small fine-tuned on our training set achieves 79.5\% accuracy (76.4\% for vanilla T5-small). 

We aim to see if using an entailment signal from a moderately good article-summary entailment classification model would help produce summaries that satisfy the AgreeSum criteria.

\section{AgreeSum Baselines and Approaches}
\label{sec:methods}

\subsection{Notations and Baselines}

We first provide some notations so as to allow easier discussion. Suppose the clusters of articles in the training set are denoted by $\{\vc_i\}_{i \in N}$. Each cluster contains at most four articles: $\vc_i = \{\vx_i^{(j)}\}_{j \leq 4}$, with the summary $\vy_i$. 

For cluster $i$, let $\ve_i$ be the set of indices that correspond to articles that entail the summary. For example, in Figure~\ref{fig:data-example} which corresponds to the $i$th cluster, we have $\ve_i = \{1,3\}$. 
Let $E$ and $D$ denote the encoder and the decoder of PEGASUS, respectively. 

The following baselines are based on PEGASUS pretrained using gap-sentence prediction on HugeNews with 1.5 billion articles (3.8 terabytes). 
{Note} that despite being an SDS model, PEGASUS also achieves near-SoTA results on the Multi-News MDS dataset \citep{fabbri-etal-2019-multi}, so it is a competitive baseline for MDS as well.

\paragraph{B1: finetuning on $(\vx^{(0)}_i, \vy_i)$ pairs.} Recall that for each summary $\vy_i$, we drop the WCEP-linked article $\vx^{(0)}_i$ to prevent excessive textual overlaps between the input cluster and the output summary. In this baseline, however, we use $\vy_i$'s in the training set and its corresponding $\vx^{(0)}_i$'s (5452 pairs) for supervised finetuning. 

Why can we use $\vx^{(0)}_i$ as gold targets? On WCEP, it is reasonable to assume that the summaries are directly connected to the linked articles, and therefore, $\vx^{(0)}_i$ entails $\vy_i$. Moreover, these summaries also inform the model of the style of the WCEP summaries. However, the downside is that the model could potentially prioritize extractions from the articles over entailment (i.e., the property that the generated summary is entailed in each article), as we would see in Section~\ref{sec:results}.

\paragraph{B2: concatenating truncated inputs.} We fine-tune PEGASUS on the following: for each cluster $\vc_i$, we truncate $\{\vx_i^{(j)}\}_j$ for each $j \in \ve_i$ and concatenate them such that the concatenated sequence has length $\leq 1024$, given hardware constraints. We use a special symbol to delineate article boundaries.

\paragraph{B3: B1+B2.} We first train using B1 and then fine-tune using B2, which may improve over B1 or B2 alone.

\paragraph{B4: merging encodings and decode.} Inspired by \citet{chu2019meansum}, we first encode $\vx_i^{(j)}$ separately for each $j \in \ve_i$, and pass the average of the encodings to the decoder (i.e., $\frac{1}{|\ve_i|} \sum_{j \in \ve_i} E(\vx_i^{(j)})$). Next, we do supervised learning based on the WCEP summaries.

\paragraph{B5: best lead-1 sentence by entailment score.} We first extract the first sentence of each article in the cluster. Next, we rerank the sentences using an entailment score, which is the mean of the binary entailment labels, predicted by fine-tuned T5-large (Section~\ref{sec:entailment-dataset}), between each article in the cluster and the given sentence. We choose the summary corresponding to the highest score, breaking ties randomly.

\subsection{AgreeSum Model (ASM): Leveraging Unannotated Data}

We introduce another baseline model called ASM. The distinguishing feature is that to target cluster-summary entailment (meaning the summary is entailed in each article in the cluster), we attempt to use the T5-entailment-classification results (discussed in Section~\ref{sec:entailment-dataset}) as a training signal. 

\paragraph{Supervised pretraining and training.}

Given pretrained PEGASUS, we first fine-tune according to B1. Next, for each cluster, we concatenate all elements of $\vc_i$ similar to B2, and mask out the articles that do not entail $\vy_i$ by the padding symbol; effectively, we use $\{\vx_i^{(j)}\}_{j \in \ve_i}$ as input to the transformer. Then, we use the standard cross-entropy loss $L_{ce}$ to train the summarizer. 
However, only a small number of clusters are annotated, providing a very limited supervised signal.

\paragraph{Unsupervised entailment loss.}

We complement $L_{ce}$ with the entailment loss $L_e$ to learn entailment behavior. 
We fine-tune T5-small on our training dataset to predict entailment, and use this as our entailment classifier $F_e$. In practice, we obtain T5-outputs using remote procedure calls (RPCs). 

$F_e$ takes in an article $\vx$ and a summary $\vy$ as inputs, and outputs $1$ if $\vx$ entails $\vy$ and $-1$ otherwise. 
The loss $L_e$ is based on policy gradients \citep{williams1992simple,sutton2000policy}; we aim to maximize a sequence-level metric of a summary {decoded from the model during training}: 
$\mathbb{E}_{\tilde{\vy} \sim p_{\phi}} \sum_{k=1}^4 F_e(\vx_i^{(k)}, \tilde{\vy})$ where $
\phi$ stands for the parameters of the encoder $E$ and the decoder $D$.
We thus have the following gradient: 
$$\nabla_\theta L_e(\phi) = - \mathbb{E}_{\tilde{\vy} \sim p_{\phi}} \nabla_\phi \log p_{\phi}(\tilde{\vy}) \hat{Q}(\tilde{\vy}),$$ where $$\hat{Q}(\tilde{\vy}) = \sum_{k=1}^4 F_e(\vx_i^{(k)}, \tilde{\vy})$$ is the sequence-level return. 
Intuitively, during training, we sample a summary $\tilde{\vy}$ from our model, and run it through the T5 entailment classifier (which is separate from our summarizer) to obtain $\hat{Q} (\tilde{\vy})$. We then weight the MLE gradient (taking $\tilde{\vy}$ as the target) by $\hat{Q} (\tilde{\vy})$. $L_e$ thus aims to guide our summarizer to generate summaries that entail in all or most of the articles.

\paragraph{Unsupervised language loss.} 

Only using the entailment loss $L_e$ may result in degenerations. One way to encourage the model to generate fluent summaries and summaries that look like WCEP summaries is by using 
GAN-style \citep{goodfellow2014generative} objectives, which have achieved good performance in some conditional generation tasks like textual style transfer and machine translation \citep{shen2017style,wu18adversarial,pang-gimpel-2019-unsupervised}. 
We thus use $F_l$, a ``language classifier'' (i.e., discriminator) that distinguishes model generations from real dataset summaries (no matter annotated or not), to force our summarizer to generate sentences that look like the human-written summaries. 
$F_l$ is based on the CNN settings by \citet{kim-2014-convolutional} identical to the setting introduced in Section~\ref{sec:entailment-dataset}, while the inputs to CNNs are sentence representations obtained using PEGASUS. 

Specifically, suppose there are $k$ examples in a minibatch, then 
\begin{align}
    L_l = - \frac{1}{k} \sum_{i=1}^k \left[ \log F_l(\vh_i) + \log (1 - F_l(\tilde{\vh}_i)) \right], \nonumber
\end{align}
where $\vh_i$ is the decoder hidden states of WCEP summaries, and $\tilde{\vh}_i$ is the decoder hidden states of model generations, by professor forcing \citep{lamb2016professor}. In the professor forcing algorithm, the input to $F_l$ is the hidden states instead of hard tokens so as to address the mismatch between training-time sequence prefix and test-time sequence prefix.

\paragraph{Summary.}
We alternate updates among the following:  
\begin{enumerate}
    \item [(1)] supervised training: updating $E$, $D$ to minimize the loss $L_{ce}$;
    \item [(2)] unsupervised training: updating $E$, $D$ to minimize the loss $L_{e}- \lambda L_l$;
    \item [(3)] language classifier training: updating $F_l$ to minimize the loss $L_l$.
\end{enumerate}

\begin{table*}[h!]
\setlength{\tabcolsep}{3pt}
\centering
\resizebox{0.995\textwidth}{!}{
\begin{tabular}{lcccccccccccccccc}
\toprule

\noalign{\smallskip}

& & \multicolumn{3}{c}{ROUGE-dev} & \multicolumn{3}{c}{ROUGE-test$^{\ddagger}$} & \multirow{2}{*}{\makecell[c]{article-summary \\ entail \% ($\uparrow$) $^\mathparagraph$}} & \multirow{2}{*}{\makecell[c]{cluster-summary \\ entail \% ($\uparrow$) $^\mathparagraph$}} & \multirow{2}{*}{\makecell[c]{hallucination \\ \% ($\downarrow$) $^\mathparagraph$}} & \multirow{2}{*}{\makecell[c]{language \\ \% ($\uparrow$)}} & \multicolumn{4}{c}{$n$-gram overlap \% ($\downarrow$)} \\ 
\cline{3-5} \cline{6-8} \cline{13-16}
model & & 1 & 2 & L & 1 & 2 & L & & & & & $n=3$ & $n=4$ & $n=5$ & $n=6$ \\
\midrule
B1 & & 45.7 & \cellcolor{cyan!30}25.0 & \cellcolor{cyan!30}38.3 & \cellcolor{cyan!30}30.7 & \cellcolor{cyan!30}12.3 & \cellcolor{cyan!30}24.2 & 56.8 / 54.0 & \cellcolor{red!10}22.7 / 20.1 & 6.0 / 6.0 & \cellcolor{red!10}90.0 & 60.4 & 49.8 & 42.6 & 38.1 \\ 
B2 & & 46.1 & 23.5 & 37.9 & 28.5 & 10.4 & 22.0 & \cellcolor{red!10}50.8 / 49.0 & 23.3 / 22.2 & 18.7 / 19.5 & 92.0 & 27.8 & 13.5 & 7.3 & 4.1 \\ 
B3 & & \cellcolor{cyan!30}47.1 & 23.7 & 38.1 & 29.1 & 10.8 & 22.4 & 49.7 / 50.7 & 22.7 / 23.3 & \cellcolor{red!10}16.0 / 20.0 & \cellcolor{cyan!30}96.0 & \cellcolor{cyan!30}26.1 & \cellcolor{cyan!30}12.8 & \cellcolor{cyan!30}6.9 & \cellcolor{cyan!30}4.0 \\ 
B5 & & \cellcolor{red!10}33.9 & \cellcolor{red!10}13.6 & \cellcolor{red!10}26.1 & \cellcolor{red!10}7.70 & \cellcolor{red!10}2.59 & \cellcolor{red!10}5.39 & 58.8 / 65.1 & 22.0 / 27.3 & \cellcolor{cyan!30}0.0 / 0.0 & 93.3 & \cellcolor{red!10}100.0 & \cellcolor{red!10}100.0 & \cellcolor{red!10}100.0 & \cellcolor{red!10}100.0 \\
ASM & & 44.4 & 22.8 & 37.0 & 27.5 & 11.0 & 22.4 & \cellcolor{cyan!30}62.8 / 66.0 & \cellcolor{cyan!30}30.7 / 39.1 & 10.0 / 8.8 & \cellcolor{cyan!30}96.0 & 47.3 & 36.4 & 30.2 & 26.1 \\ 
\bottomrule
\end{tabular}
}
\caption{Results (on test set if not specified). 
$^\mathparagraph$: the first number corresponds to T5-large-evaluated results, and the second number corresponds to human-evaluated results. For each row, human raters annotated the entire test set; each article-generation pair is annotated by three raters, and we take the majority answer. The best result in each column is in blue; the worst in red. $n$-gram overlap: the proportion of generation $n$-grams that are also in the source. $^\ddagger$: the test set WCEP summaries should not be treated as references, given that there is no guarantee that the WCEP summary is entailed in each of the articles; the test set WCEP summaries are provided as an approximate measure of informedness. Note: B4 results in extensive hallucinations and very low ROUGE ($\sim$10), and the ablation study of ASM without $L_l$ produces heavy degenerations, so they are omitted. 
\label{tab:results}} 
\end{table*}

\subsection{Implementation}
\label{sec:implementation}

We implement our model as a fork of the open-source 568M-parameter PEGASUS model \citep{zhang2020pegasus}.\footnote{\url{https://github.com/google-research/pegasus}} 
We initialize PEGASUS from the ``mixed \& stochastic'' checkpoint, which was pre-trained for 1.5M steps. 

All baseline models are fine-tuned with a learning rate of 1e-4. All of the ASM models use 5e-5 (tuned in \{1e-5, 5e-5, 1e-4, 2e-4\}). All models use beam search for decoding (beam size 8, beam alpha 0.8). Given hardware constraints, all models use a max input length of 1024. The max output decoding length is set to be 128. In addition, we tune $\lambda \in \{0.05, 0.1, 0.3, 0.5, 1, 2\}$ and choose $\lambda=0.1$ for all reported experiments that include the $L_l$. 
We do not change any other default hyperparameter settings adopted from PEGASUS.
Please refer to the appendix for more details.

\section{Results}
\label{sec:results}

The following two models are not included in the table given their poor performance. 
(1) The ablation study of ASM without language loss does not produce meaningful outputs. In this case, given that the T5 entailment classifier does not encourage language quality, the summaries in fact degenerate heavily. 
(2) Model B4 (merging encodings and decode) is omitted from the table given very poor ROUGE performance ($\sim$10) and extensive hallucination. Our conjecture is that the mean of the encodings of articles  
does not correspond to a meaningful encoding in the case of PEGASUS.

\subsection{Agreement and Hallucination}
\label{sec:agreement-eval}

First, we claim that ASM achieves {better agreement}. There are two types of agreement: \emph{article-summary} agreement which is the proportion of summaries entailed in the articles, and \emph{cluster-summary} agreement which is the proportion of clusters in which \emph{all} articles entail the summary. Table~\ref{tab:results} reports both the T5-automatic evaluation and the human evaluation results on agreement. 

\paragraph{(Preliminary) automatic metrics.}

Given an article-summary pair, we use T5-large fine-tuned on our entailment dataset (Section~\ref{sec:entailment-dataset}) to predict whether the summary is entailed in the article. For article-summary agreement, we compute the percentage of ``entailed'' classifications. For cluster-summary agreement, we compute the percentage of clusters where all article-summary pairs lead to ``entailed'' classifications. 
We see that the T5-evaluation results for ASM models perform better than the respective results for baseline models.

\paragraph{Human evaluation.}

For each row of Table~\ref{tab:results}, human raters annotated the entire test set (150 clusters, which corresponds to 600 article-generation pairs), on whether the generated summaries are entailed in the article. Workers were asked: 
\begin{quote}
    \textit{Does the article contain all the information presented in the summary?} 
\end{quote}The full prompt is available in the appendix. One design choice is that we merge all article-summary pairs for each cluster together into one task/HIT. Therefore, each task/HIT corresponds to four article-summary annotations. 
To reduce the inherent variance in human evaluation, each article-generation pair is annotated by {three} different raters, and we take the majority answer. Please see the appendix for more details.

For article-summary agreement, we see that the human evaluation for the ASM models performs a little better than the extractive results (B5). ASM models perform more than 10 points better than the best abstractive baseline results (B1), which is in turn $\sim$5 points better than the other abstractive baseline results (B2, B3). For cluster-summary agreement, this improvement is even clearer, with ASM models performing more than 10 points better than any baseline. 

In addition, we see that ASM {reduces hallucination} compared to B2 and B3. One way to approximate hallucination is by the number of clusters in which none of the articles entail the generated summary. Using both automatic and human evaluation results, we see that our model does better than B2 and B3, but a little worse than B1 which copies extensively from the source articles.

\begin{table*}[h!]
\setlength{\tabcolsep}{3pt}
\centering
\resizebox{0.995\textwidth}{!}{
\begin{tabular}{lcccccccccccccccc}
\toprule

\noalign{\smallskip}
& & \multicolumn{3}{c}{ROUGE-dev} & \multicolumn{3}{c}{ROUGE-test$^\ddagger$} & \multirow{2}{*}{\makecell[c]{article-summary \\ entail \% ($\uparrow$) $^\mathparagraph$}} & \multirow{2}{*}{\makecell[c]{cluster-summary \\ entail \% ($\uparrow$) $^\mathparagraph$}} & \multirow{2}{*}{\makecell[c]{hallucination \\ \% ($\downarrow$) $^\mathparagraph$}} & \multirow{2}{*}{\makecell[c]{language \\ \% ($\uparrow$)}} & \multicolumn{4}{c}{$n$-gram overlap \% ($\downarrow$)} \\ 
\cline{3-5} \cline{6-8} \cline{13-16}
model & & 1 & 2 & L & 1 & 2 & L & & & & & $3$ & $4$ & $5$ & $6$ \\
\midrule
B1-entdec8 & & 45.7 & \cellcolor{cyan!30}24.8 & \cellcolor{cyan!30}38.3 & \cellcolor{cyan!30}29.5 & \cellcolor{cyan!30}12.0 & \cellcolor{cyan!30}23.4 & 60.8 / 56.8 & 26.0 / 26.2 & 4.67 / 6.0 & 93.3 & 61.1 & 50.7 & 43.5 & 38.8 \\ 
B2-entdec8 & & 46.1 & 23.3 & 38.0 & 28.5 & 10.4 & 22.0 & 53.2 / 58.3 & 23.3 / 34.2 & 18.0 / 18.1 & 95.3 & 28.1 & 13.8 & 7.4 & 4.1 \\
B3-entdec8 & & \cellcolor{cyan!30}47.2 & 23.8 & 38.1 & 28.9 & 10.6 & 22.1 & 53.7 / 59.3 & 24.7 / 33.6 & 15.3 / 13.4 & 94.0 & \cellcolor{cyan!30}26.7 & \cellcolor{cyan!30}13.0 & \cellcolor{cyan!30}6.9 & \cellcolor{cyan!30}4.0 \\
ASM & & 44.4 & 22.8 & 37.0 & 27.5 & 11.0 & 22.4 & \cellcolor{cyan!30}62.8 / 66.0 & 30.7 / 39.1 & \cellcolor{cyan!30}10.0 / 8.8 & \cellcolor{cyan!30}96.0 & 47.3 & 36.4 & 30.2 & 26.1 \\ 
ASM-entdec8 & & 44.8 & 23.4 & 37.7 & 26.9 & 11.0 & 22.1 & 68.2 / 63.3 & 40.0 / 37.0 & 8.0 / 10.9 & 91.3 & 46.9 & 35.4 & 29.1 & 24.6 \\ 
ASM-entdec16 & & 44.7 & 23.1 & 37.5 & 26.3 & 10.8 & 21.5 & 70.5 / 63.8 & \cellcolor{cyan!30}42.7 / 39.9 & 8.0 / 11.9 & 90.0 & 48.7 & 36.9 & 30.3 & 25.6 \\ 
\bottomrule
\end{tabular}
}
\caption{Results using the entdec decoding strategy. The results correspond to the test set performance, if not specified. 
$^\mathparagraph$: the first number in each cell corresponds to T5-evaluated results, and the second corresponds to human-evaluated results. Given that B5 relies on pure extraction, the entdec methods are not applicable. $^\ddagger$: the test set WCEP summaries should not be treated as references, given that there is no guarantee that the WCEP summary is entailed in each of the articles; the test set WCEP summaries are provided as an approximate measure of informedness.
\label{tab:results-dec}} 
\end{table*}

\subsection{Discussion on ROUGE}
\label{sec:rouge-eval}

In SDS tasks, \citet{durmus-etal-2020-feqa} and \citet{wang-etal-2020-asking} observe that ROUGE and BERTScore have a small correlation with summary factuality. 

A hallucination or non-entailment\footnote{Non-entailed summaries are not necessarily hallucinations, given that the non-entailed summaries could correspond to some articles in the cluster but not the rest of the articles.} can have major text-span overlaps with the reference, thereby having a large ROUGE score. 
Given the nature of AgreeSum, and by Table~\ref{tab:results}, we confirm that high ROUGE does not imply entailment and should not be considered heavily in evaluation. 

On the other hand, intuitively, we do recognize that an overly small ROUGE may indicate bad generations (e.g., extremely short generations, off-topic generations, and other degenerations like repetitions), which is the case for B4 generations as well as ASM-minus-$L_l$ generations. 

Thus, practitioners need to rely on and determine the desired tradeoff between the following two automatic metrics: (1) ROUGE as a coarse proxy for summary quality and informedness, and (2) entailment-related and hallucination-related metrics (Section~\ref{sec:agreement-eval}). 

As a reminder, the development set summaries can be treated as gold-standard summaries. However, the test set summaries are not gold-standard summaries; they are only provided so as to allow one way to measure generation quality and informedness. Unlike the development set, none of the test-set input articles are filtered out through the summary-article entailment annotation procedure, given that we do not want to introduce potential bias through too much manipulation and filtering on raw test clusters.

\subsection{More Observations}

\paragraph{Language.} We also asked workers to judge the language of the generations: 
\begin{quote}
    \textit{Is this summary coherent and well-written with no self-contradictions or capitalization, spelling, punctuation, or grammar errors?}
\end{quote}
The full prompt is in the appendix.
We see that the ASM-generated summaries are marginally better at the above. However, ASM without language loss results in heavy degenerations. 
On a separate note, we see that ASM generations tend to copy more than B2 and B3, but less than B1. 

\paragraph{Examples.}

The appendix contains a few examples that compare generations from different models. For example, in Table~\ref{tab:ex-1}, given four articles, we see the different generations that the systems produce. Article 4 is an opinion piece. ASM model correctly abstractively summarizes Article 2 and 4 in a way that agrees with 1 and 3. B1 copies from Article 1, while B2 and B3 have hallucinations. Given space constraints, please refer to the appendix.

\subsection{Extension: Post-hoc Entailment Reranking after Decoding a Beam}

To generate summaries that achieve better agreement, we also attempted a decoding trick which we name as entailment-oriented decoding, denoted by {entdec} in Table~\ref{tab:results-dec}. 

We first define an entailment score used in this case. Given a cluster of articles and a generated summary, the entailment score is the mean of the T5-large-predicted binary labels (1 corresponds to ``entailed'' and 0 corresponds to ``not entailed''). 

A model is suffixed as X-entdec$k$ if we decode from X using beam search with beam size $k$; and after obtaining the size-$k$ beam, we select the generation that corresponds to the largest T5 entailment score. We pick the beam with the largest score, using the original beam probabilities to break ties. 
Intuitively, this trick picks the best-entailment summary locally given that the generations in the same beam are usually similar. 

Table~\ref{tab:results-dec} shows that the entdec-trick generations indeed achieve higher article-summary agreement and human-summary agreement. We see that ASM still maintains the advantage in agreement, even if compared to entdec-decoded generations from other baselines.

\subsection{Discussion: Improving Entailment Using T5-Based NLI-Style Models}

\citet{falke-etal-2019-ranking} find that NLI models trained on standard NLI datasets do not offer robust benefits to improving summarization factuality. \citet{maynez-etal-2020-faithfulness}, on the other hand, rerank four summaries generated by four different models using BERT-based MNLI models, and find small improvements in faithfulness and factuality. However, these works rank different summaries after decoding assuming an existing summarizer (similar to our entdec trick), instead of updating the model parameters directly. 
Our contribution lies in the fact that we successfully use entailment models to improve the model during training time.

The major feature of our T5-based NLI model is that (1) our NLI model is based on multi-task-pretrained T5, implying that pretrained T5 can already handle article-length inputs well in certain tasks, and (2) our model is obtained after finetuning on our article-summary entailment dataset. Therefore, our T5-based NLI model is much better adjusted to the length of the premises (given that traditional NLI tasks correspond to sentence-level entailment, but our case corresponds to article-summary entailment). 
We thus see that using simple T5-based binary signals can successfully improve entailment. However, more complicated modeling may be necessary if the AgreeSum cluster size becomes much larger.

\section{Conclusion}
\label{sec:conclusion}

We discuss the AgreeSum task with its dataset, and a range of baseline models. AgreeSum is timely given the recent focus on summarization faithfulness. In fact, we show that the summaries produced by several powerful pretraining-based baseline models are not able to follow AgreeSum's requirements satisfactorily. We welcome the community to contribute more advanced methods that work well on AgreeSum, especially when only a small subset of the dataset is labeled with article-summary entailment information. 

Within the AgreeSum dataset, we also provide article-summary entailment annotations on a subset of clusters, which we hope can contribute to the recent effort in improving abstractive summarization faithfulness. 

Moreover, while there is contemporaneous development of complex approaches to encourage generated abstractive summaries to be entailed in the source articles, we show that it is feasible to improve the entailment behavior of generated summaries based on a binary article-summary entailment classifier.

\section*{Acknowledgement}

The authors would like to thank Pepa Atanasova, He He, Abe Ittycheriah, Jialu Liu, Tianqi Liu, Ji Ma, Shashi Narayan, Alicia Parrish, Jiaming Shen, Gonçalo Simões, and You (Will) Wu (alphabetical order) for the valuable discussions, and the anonymous reviewers for the detailed reviews.

\bibliographystyle{acl_natbib}
\bibliography{anthology,acl2021}

\clearpage

\appendix

\section{Appendix}
\label{sec:appendix}

\subsection{Human Evaluation}
\label{app:human-eval}

Recall that human evaluation results are reported in Table~\ref{tab:results} and Table~\ref{tab:results-dec}. Specifically, we asked human raters to annotate the entailment relationship of each article-generation pair. Each row in the table corresponds to a model, and for each model, human raters annotated the entire test set that contains 150 clusters (which corresponds to 600 article-generation pairs). Moreover, to reduce the variance of human evaluation and to improve the confidence to our claims, each article-generation pair is annotated by three different raters, and we take the majority answer. 

Our prompts to human raters are described as follows. We merge all article-summary pairs for each cluster together into one task/HIT. Therefore, each task/HIT corresponds to four article-summary annotations.

\textbf{Instruction at the top of the annotation page:} ``In this task you will be given one summary of a news story and the text of four news articles. You will be asked to evaluate whether the summary is coherent and well written with no self-contradictions or capitalization, spelling, punctuation, or grammatical errors.
Furthermore, for each news article, you will be asked to evaluate: whether the article contains ALL the information presented in the summary. This is equivalent to asking, `Using only the text of this news article, would it be possible for someone to write this summary?'''

\textbf{Language:} ``Is this summary coherent and well-written with no self-contradictions or capitalization, spelling, punctuation or grammar errors? Select ``Yes" if this is a coherent summary written in fluent English with perfect grammar and style. Select “No” if the summary contains one or more capitalization, spelling, punctuation, or grammatical errors, or is incoherent, self-contradictory, or otherwise badly written.''

\textbf{Entailment:} ``Does the article contain all the information presented in the summary?''

\subsection{Entailment Annotation in Dataset Creation}

We used the same entailment prompt described in Section~\ref{app:human-eval} to obtain entailment information for a subset of the clusters. The difference is that we used five annotators per article-summary pair instead of three, to ensure the quality of the supervised split of the training set as well as the entire development set.

\subsection{More on Reproducibility}

We implement our model as a fork of the open-source 568M parameter PEGASUS model.\footnote{\url{https://github.com/google-research/pegasus}}

Baseline models were trained for 20k steps, with the best checkpoint selected based on dev-R2. Final selections are B1 at 3k steps, B2 at 6k steps, and B3 at 4k steps. Proposed model was trained until convergence based on T5-small entailment scores reported during the sampling steps in the policy gradients REINFORCE algorithm. Final selection of this model was at 43k steps.

\paragraph{Runtime details.}
Given the chosen PEGASUS implementation, our models are trained on TPUs. Given hardware constraints, $F_e$, implemented via T5-small fine-tuned on our annotated entailment dataset, is served on CPU on a separate machine. In fact, we run 80 replicas of this machine to improve throughput. We make RPC calls \citep{nelson1981remote} to this cluster of T5-serving machines during policy gradients training (using the f.contrib.rpc module). Baseline models took approximately 24 hours each to train, with the proposed model taking approximately 4 days. Our conjecture is that for the latter, most of the time is spent on the communication between PEGASUS and T5, as well as the expensive computation of T5 on CPU.

\subsection{Examples}

We now provide some example generations of the AgreeSum task. Given that the input articles are long, we comment out parts of the articles in the tables. Please refer to Table~\ref{tab:ex-2}, Table~\ref{tab:ex-3}, and their captions.

For example, in Table~\ref{tab:ex-1}, Article 4 is an opinion piece. ASM model correctly abstractively summarizes Article 2 and 4 in a way that agrees with 1 and 3. B1 copies from Article 1, while B2 and B3 have hallucinations. In Table~\ref{tab:ex-2}, all models except for B1 perform well here. B1 entails Article 1 and 2, but not 3 and 4. Similar comparisons can be made in Table~\ref{tab:ex-3}.

\begin{center}
\begin{table*}[h]
\small
\begin{tabular}{p{15.5cm}}
\midrule \textbf{ASM:} Sérgio Moro resigns as Brazil's Justice Minister after accusing President Jair Bolsonaro of interfering in the country's federal police.\\ 
\midrule \textbf{B1:} Brazil's Supreme Court authorises a police investigation into President Jair Bolsonaro. \\ 
\midrule \textbf{B1+entdec8:} Brazil's Supreme Court authorises a police investigation into President Jair Bolsonaro. \\ 
\midrule \textbf{B2:} The Minister of Justice of Brazil, Sérgio Moro, resigns after accusing President Jair Bolsonaro of interfering in the operations of the federal police. \\ 
\midrule \textbf{B2+entdec8:} The Minister of Justice of Brazil, Sérgio Moro, resigns after accusing President Jair Bolsonaro of interfering in the operations of the federal police.\\ 
\midrule \textbf{B3:} The Minister of Justice of Brazil, Sérgio Moro, resigns after accusing President Jair Bolsonaro of interference in the federal police.\\ 
\midrule \textbf{B3+entdec8:} The Minister of Justice of Brazil, Sérgio Moro, resigns after accusing President Jair Bolsonaro of interference in the federal police.\\
\midrule \textbf{B5:} Brazil's government has been plunged into turmoil after the resignation of one of Jair Bolsonaro's most powerful ministers sparked protests, calls for the president's impeachment and an investigation into claims he had improperly interfered in the country's federal police. \\
\midrule \textbf{Article 1:} [Link: https://www.theguardian.com/world/2020/apr/24/justice-ministers-sacking-plunges-brazil-into-turmoil; the article is not duplicated due to length and copyright] \\
\midrule \textbf{Article 2:} [Link: https://www.ft.com/content/62d04bb5-6825-41ec-b263-4ceeaec58049; the article is not duplicated due to length and copyright] \\
\midrule \textbf{Article 3:} [Link: https://www.theguardian.com/world/2020/apr/26/bolsonaro-in-fresh-crisis-over-sons-alleged-links-to-fake-news-racket; the article is not duplicated due to length and copyright] \\
\midrule \textbf{Article 4:} [Link: https://www.thedailybeast.com/brazils-justice-minister-sergio-moro-quits-accuses-president-jair-bolsonaro-of-misconduct-resigns; the article is not duplicated due to length and copyright] \\
\bottomrule
\end{tabular}
\caption{Example generations from test set. All models except for B1 perform well here. B1 entails Article 1 and 2, but not 3 and 4.
\label{tab:ex-2}}
\end{table*}
\end{center}

\begin{table*}[h!]
\footnotesize
\begin{tabular}{p{15.5cm}}
\midrule \textbf{ASM:} The NFL announces that it will play during the coronavirus pandemic. \\ 
\midrule \textbf{B1:} NFL Commissioner Roger Goodell sends a letter to fans outlining the league's plans to play during the coronavirus pandemic. \\ 
\midrule \textbf{B2:} The NFL cancels the remainder of the 2020 season due to the coronavirus outbreak. \\ 
\midrule \textbf{B3:} The NFL cancels the remainder of the 2019 preseason due to the ongoing coronavirus outbreak. \\ 
\midrule \textbf{B5:} NEW YORK (AP) — NFL Commissioner Roger Goodell has sent a letter to fans outlining the league's plans to play during the coronavirus pandemic. \\
\midrule \textbf{Article 1:} [Link: https://apnews.com/article/nfl-sports-virus-outbreak-health-football-c10944a1b88bd593198b660d207c7b56; the article is not duplicated due to length and copyright] \\
\midrule \textbf{Article 2:} [Link: https://www.washingtonpost.com/sports/2020/07/27/nfl-cautiously-optimistic-despite-mlb-coronavirus-outbreak/; the article is not duplicated due to length and copyright] \\ 

\midrule \textbf{Article 3:} [Link: https://www.wxyz.com/sports/roger-goodell-sends-letter-to-nfl-fans-explaining-plans-for-season; the article is not duplicated due to length and copyright] \\
\midrule \textbf{Article 4:} [Link: https://www.wsj.com/articles/nfl-playersand-a-lot-of-new-england-patriotsare-opting-out-of-the-2020-season-11596030916; the article is not duplicated due to length and copyright] \\ 
\bottomrule
\end{tabular}
\caption{Example generations from test set. Note in particular that Article 4 is an opinion piece. The ASM model correctly abstractively summarizes Article 2 and 4 in a way that agrees with 1 and 3. B1 and B5 copy from Article 1, while B2 and B3 have hallucinations. More examples can be found in the appendix. 
\label{tab:ex-1}
}
\end{table*}

\begin{center}
\begin{table*}[h]
\small
\begin{tabular}{p{15.5cm}}
\midrule \textbf{ASM:} Late-night talk shows in New York City begin taping without audience members. \\ 
\midrule \textbf{B1:} The Late Show with Stephen Colbert, The Tonight Show with Jimmy Fallon, and The Full Frontal with Samantha Bee begin taping without audience members due to the spread of the coronavirus in New York City.\\ 
\midrule \textbf{B1+entdec8:} The Late Show with Stephen Colbert, The Tonight Show with Jimmy Fallon, and The Full Frontal with Samantha Bee begin taping without audience members due to the spread of the coronavirus in New York City.\\ 
\midrule \textbf{B2:} In response to the outbreak of coronavirus in the United States, all New York-based late-night talk shows will cease taping without an audience.\\ 
\midrule \textbf{B2+entdec8:} In response to the outbreak of coronavirus in the United States, all New York-based late-night talk shows will cease taping without an audience.\\ 
\midrule \textbf{B3:} Following the lead of the Centers for Disease Control and Prevention, all late-night talk shows in New York City will stop taping and start without an audience due to the outbreak of coronavirus. all late-night shows in New York City will stop taping and start without an audience Due to the outbreak of coronavirus, all late-night shows in New York City will stop taping and start without an audience.\\ 
\midrule \textbf{B3+entdec8:} Following the lead of the Centers for Disease Control and Prevention, all late-night talk shows in New York City will stop taping and start without an audience due to the outbreak of coronavirus. all late-night shows in New York City will stop taping and start without an audience Due to the outbreak of coronavirus, all late-night talk shows in New York City will stop taping and start without an audience Due to the outbreak of coronavirus, all late-night talk shows in New York City will stop filming and start without an audience due to the outbreak of coronavirus.\\
\midrule \textbf{B5:} The New York late-night circuit was the antithesis of “Live in Front of a Studio Audience” this week, prerecording shows without a crowd due to coronavirus fears. \\
\midrule \textbf{Article 1:} [Link: https://www.latimes.com/entertainment-arts/tv/story/2020-03-13/coronavirus-jimmy-fallon-stephen-colbert-no-audience; the article is not duplicated due to length and copyright] \\
\midrule \textbf{Article 2:}
[Link: https://variety.com/2020/tv/news/late-night-shows-new-york-coronavirus-1203530972/; the article is not duplicated due to length and copyright] \\
\midrule \textbf{Article 3:} [Link: ]https://abcnews.go.com/Entertainment/wireStory/late-night-comics-adjust-shows-audience-69581648; the article is not duplicated due to length and copyright] \\
\midrule \textbf{Article 4:} [Link: https://apnews.com/article/38f1afde2a5676fd3c2377f3719f5c86; the article is not duplicated due to length and copyright] \\
\bottomrule
\end{tabular}
\caption{Example generations from test set. ASM is very abstractive and entails all articles. B1 only entails Article 1 and 2. B2 incorrectly says that shows will ``cease taping.'' B3 has repetition issues.
\label{tab:ex-3}}
\end{table*}
\end{center}

\end{document}